\documentclass{article}
\usepackage{spconf,amsmath,graphicx}
\usepackage{booktabs}
\usepackage{hyperref}
\usepackage{soul}

\title{Distilling Knowledge from Ensembles of Acoustic Models \\ for Joint CTC-Attention End-to-End Speech Recognition}
\name{Yan Gao$^1$, Titouan Parcollet$^{2,1}$, Nicholas D. Lane$^{1,3}$}
\address{$^1$University of Cambridge, United Kingdom, 
  $^2$Avignon University, France \\
  $^3$Samsung AI, Cambridge, United-Kingdom}
\begin{document}
%
\maketitle
\begin{abstract}
Knowledge distillation has been widely used to compress existing deep learning models while preserving the performance on a wide range of applications. In the specific context of Automatic Speech Recognition (ASR), distillation from ensembles of acoustic models has recently shown promising results in increasing recognition performance. In this paper, we propose an extension of multi-teacher distillation methods to joint CTC-attention end-to-end ASR systems. We also introduce three novel distillation strategies. The core intuition behind them is to integrate the error rate metric to the teacher selection rather than solely focusing on the observed losses. In this way, we directly distill and optimize the student toward the relevant metric for speech recognition. We evaluate these strategies under a selection of training procedures on different datasets (TIMIT, Librispeech, Common Voice) and various languages (English, French, Italian). In particular, state-of-the-art error rates are reported on the Common Voice French, Italian and TIMIT datasets. 

\end{abstract}
\section{Introduction}
\label{sec:intro}

Knowledge Distillation (KD) \cite{hinton2015distilling}, also known as teacher-student training, is commonly used to narrow the gap of performance between a smaller model and a larger one \cite{rusu2015policy, kim2016sequence,chen2017learning,mishra2017apprentice,wang2019private}. A typical KD training procedure consists of two stages. First, a deep neural network referred as the \textit{teacher} is trained in line with standard supervised training rules based on numerous samples and their corresponding ground truth labels. Second, a compressed network, the \textit{student} model, is trained on a selection of original ground truths and soft targets labelled by the teacher. These soft targets are the posterior probabilities obtained from the pre-trained teacher. Knowledge distillation has been shown to be particularly efficient to reduce the student size while matching its performance to that of the teacher. Common applications include Computer Vision (CV) \cite{chen2017learning,polino2018model,liu2018multi}, Natural Language Processing \cite{cui2017knowledge,jiao2019tinybert,sun2019patient} (NLP) and Automatic Speech Recognition (ASR) \cite{chebotar2016distilling,kurata2019guiding, kim2019knowledge,takashima2018investigation}. 

\begin{figure*}
  \centering
  \includegraphics[width=0.68\linewidth]{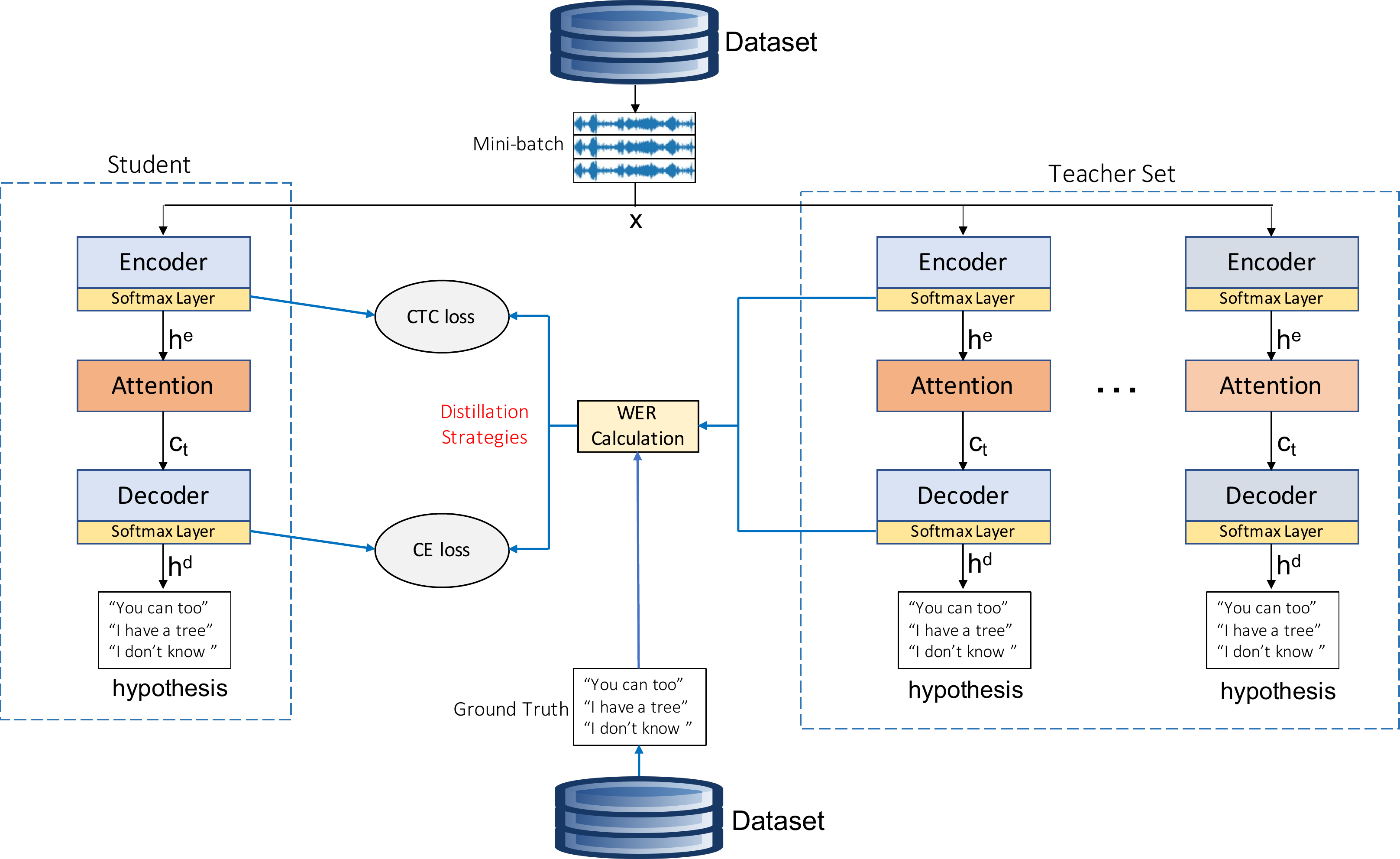}
  \caption{ Illustration of the error rate multi-teacher distillation strategies connected to a Joint CTC-Attention E2E ASR system.}
  \label{fig:speech_production}
\end{figure*}

An alternative approach to KD focuses solely on increasing the performances of the student model without considering its complexity. Distillation from ensembles of teachers has been commonly conducted under this approach. This method is referred to as the multi-teacher distillation \cite{freitag2017ensemble,fukuda2017efficient}. 

Modern deep learning based ASR systems have been shown to strongly benefit from multi-teacher distillation strategies \cite{chebotar2016distilling, fukuda2017efficient}. Empirically, ensembles of teacher models capture complementary information by making different errors that can be further distill to a student model. A critical aspect of multi-teacher distillation in the context of ASR is to find suitable strategies to maximize the distillation with respect to a specific set of teachers. For instance, \cite{chebotar2016distilling} proposed to pre-assign weights to teachers to control their impact on the distilled information. Another strategy is to sample the considered teachers randomly \cite{fukuda2017efficient}. However, both strategies may give higher weighting, and thus higher importance, to teachers that are performing worse than others in the teacher set when applied to specific sentences.

End-to-End (E2E) ASR models are particularly well suited for KD as the whole pipeline is composed of neural networks only \cite{kurata2019guiding, takashima2018investigation, huang2018knowledge}. One set of E2E ASR systems commonly rely either on the Connectionist Temporal Classification (CTC) loss \cite{graves2014towards},  Sequence to Sequence models (Seq2Seq) \cite{bahdanau2016end}, or a combination of the two \cite{kim2017joint}. While single teacher distillation to achieve acoustic model compression have been widely investigated on the CTC and Seq2Seq families of models \cite{kurata2019guiding, kim2016sequence}, works on ensembles of teachers to enhance the performances remain scarce. 

The multi-teacher setup holds untapped potential as different E2E ASR systems often lead to different transcriptions given a fixed audio sample, which strongly increases the diversity of the teachable distributions that could be distilled to the student. Therefore, it is of crucial interest to explore the use of diverse sets of E2E teachers to increase both the robustness and the performance of the student acoustic model. Potential use-cases include: Federated Learning (FL) \cite{dimitriadis2020federated, li2020federated} with hundreds of acoustic models being trained concurrently, thus needing a proper aggregation or distillation strategy to further reduce the error rate and training time; Production-oriented training pipelines of ASR systems relying on strong hyper-parameters tuning phases with multiple models that could be further used rather than discarded to improve the quality of the final model. 

In this paper, we first propose and investigate an extension of multi-teacher KD strategies to joint CTC-attention based ASR models. Motivated by error-weighted ensemble methods \cite{kuncheva2014combining}, we introduce three novel Error Rate (ER)-based multi-teacher distillation strategies. Indeed, common distillation strategies only consider the loss as an indicator to assess the teacher quality, while a more relevant scheme for ASR is to  optimise our student toward the transcription quality.

First, the \textit{``Weighted'‘} strategy enables the student to directly assign weights to all the teachers in the course of training based on the average observed ER on the processed mini-batch. The impact of the teachers is therefore dynamically changed between mini-batches. Then, the \textit{``Top-1'‘} strategy offers the student an option to choose a single teacher with respect to the best ER observed at the sentence level on the processed mini-batch. Finally, The \textit{``Top-k'‘} strategy allows the student to learn from the a set of best teachers that perform equally in terms of error rate on the processed mini-batch as equal ER do not necessarily imply identical transcripts.

In short, our contributions are: a. Introduce multi-teacher distillation for ER reduction on joint CTC-attention based E2E systems (Sec. \ref{sec:dis} \& Sec. \ref{sec:strat}); b. Propose three novel distillation strategies focusing on the reduction of the ER (Sec. \ref{sec:strat}); c. Compare all the models on four different datasets: TIMIT, Librispeech, Common Voice (CV) French and Italian (Sec. \ref{sec:results}). The code to replicate our results and to facilitate further research is distributed within the SpeechBrain\footnote{\url{https://speechbrain.github.io}} toolkit \cite{SB2021}. State-of-the-art error rates of $13.11$\%, $15.57$\% and $17.04$\% are reported on TIMIT and CV Italian, French respectively.

\section{Distillation for Joint CTC-Attention Speech Recognition}
\label{sec:dis}

Joint CTC-Attention E2E systems \cite{kim2017joint} combine a sequence-to-sequence (Seq2Seq) attention-based model \cite{bahdanau2016end} with the CTC loss \cite{graves2014towards}. The CTC is applied to facilitate the training of the attention decoder by directing the attention toward the correct alignment. 

A typical Seq2Seq model includes three modules: an encoder, a decoder and an attention module. The \textit{encoder} processes an input sequence $\textbf{x} = [x_1, ..., x_{T_x}] \in X$ with a length $T_x$, where $X$ is the whole set of training data, and creates a hidden latent representation $\textbf{h}^e = [h^e_1 , ..., h^e_{T_x}]$. Then the \textit{decoder} attends $\textbf{h}^e$ combined with an attention context vector $c_t$ obtained with the attention module to produce the different decoder hidden states $\textbf{h}^d = [h^d_1 , ..., h^d_{T_y}]$, with $T_y$ the length of the target sequence $\textbf{y} =[y_1, ..., y_{T_y}] \in Y$, and $Y$ the target utterances dataset composed with a vocabulary $V = [v_1, ... v_z]$ of $z$ tokens (\textit{e.g.} characters, tokens, words ...). Note that in a speech recognition scenario, the length of the original signal $T_x$ is much longer than the utterance length $T_y$ (\textit{i.e.} many to one).

The standard supervised training procedure of the joint CTC-Attention ASR pipeline is based on two different losses. First, the CTC loss is derived with respect to the prediction obtained from the encoder module of the Seq2Seq model:

\begin{equation}
\mathcal{L}_{CTC} = - \sum_S \log p(\textbf{y}|\textbf{h}^e),
\label{eq1}
\end{equation}
with $S = \{X, Y\}$ denoting the training dataset. Second, the attention-based decoder is optimized following the Cross Entropy (CE) loss:

\begin{equation}
\mathcal{L}_{CE} = - \sum_S \log p(\textbf{y}|\textbf{h}^d).
\label{eq2}
\end{equation}

Both losses are finally combined and controlled with a fixed hyperparameter $\alpha$ ($0\le \alpha \le 1$) as: 

\begin{equation}
\mathcal{L} = \alpha \mathcal{L}_{CE} + (1-\alpha)\mathcal{L}_{CTC}.
\label{eq3}
\end{equation}

\noindent\textbf{Single teacher distillation.} Different techniques have been proposed to adapt both $\mathcal{L}_{CE}$ and $\mathcal{L}_{CTC}$ to single teacher distillation \cite{kurata2019guiding, takashima2018investigation, huang2018knowledge}. Following these findings, we decided to follow frame-level and sequence-level distillations for $\mathcal{L}_{CE-KD}$ and $\mathcal{L}_{CTC-KD}$ respectively. More precisely, the former loss (\textit{i.e.} attentional decoder) will be computed by considering the soft label produced for every frame as the true output distribution (\textit{i.e.} not a one hot encoding) for the teacher, while the latter one receives the entire sequence from the teacher as one hot true labels. In this context, \cite{takashima2018investigation} proposed to increase the diversity by generating a list of \textit{N}-best true sentences with a beam-search. Therefore, $\mathcal{L}_{CE-KD}$ is computed as:

\begin{equation}
\mathcal{L}_{CE-KD} = - \sum_S \sum_{h^d_1}^{h^d_{T_y}} \sum_{v\in V} p_{tea}(v|\textbf{h}^d_{tea}) \log p_{st}(v|\textbf{h}^d_{st}),
\label{eq4}
\end{equation}
with, $\textbf{h}^e_{tea}$ and $\textbf{h}^e_{st}$ the hidden vector representations of the teacher and student encoders respectively,  and $p_{tea}(v|\textbf{h}^d_{tea})$ and $p_{st}(v|\textbf{h}^d_{st})$  the posterior probabilities estimated by the teacher and student models with respect to the label $v$. Then, $\mathcal{L}_{CTC-KD}$ is defined at the sequence-level instead as:

\begin{equation}
\mathcal{L}_{CTC-KD} = - \sum_S \sum_{n=1}^{N} p_{tea}(\hat{\textbf{y}}_n|\textbf{h}^e_{tea}) \log p_{st}(\hat{\textbf{y}}_n|\textbf{h}^e_{st}), 
\label{eq5}
\end{equation}
with $\hat{\textbf{y}}_n$ the $n$-th hypothesis from the set of \textit{N}-best hypothesis obtained via beam-search (\textit{i.e.} beam width=\textit{N}) for the teacher. Then, Eq. \ref{eq3} is extended to knowledge distillation:

\begin{equation}
\mathcal{L}_{KD} = \alpha \mathcal{L}_{CE-KD} + (1-\alpha)\mathcal{L}_{CTC-KD}.
\label{eq7}
\end{equation}

Finally, an extra-layer of customization is added with a global loss combining KD and supervised training: 

\begin{equation}
\mathcal{L}_{total} = \beta \mathcal{L}_{KD} + (1-\beta)\mathcal{L},
\label{eq8}
\end{equation}
with $\beta \in [0, 1]$ an hyperparameter controlling the impact of KD during the training (\textit{e.g.} $\beta=1$ in all our experiments) .

\section{Multi-teacher Error Rate Distillation}
\label{sec:strat}

Different E2E ASR models make different mistakes while transcribing the same audio recording. Therefore, distillation from multiple pre-trained teachers has potential to help the student model to improve considerably. Finding a good teacher weight assignment strategy, however, is not trivial. An existing approach \cite{chebotar2016distilling,fukuda2017efficient} is to simply compute an average over the set of teachers: 

\begin{equation}
\mathcal{L}_{multi} = \sum_m w_m \mathcal{L}_m,
\label{eq9}
\end{equation}
with $w_m \in [0, 1]$ the pre-assigned weight corresponding to the $m$-th teacher model and equal to $1/M$. $M$ is the total number of teachers composing the ensemble. However, this method gives to a poor teacher the same importance as a good one while a natural solution would be to associate well-performing teachers with higher weights.

We propose to consider the error rate metric as a proxy to determine which teacher loss to consider during distillation. Indeed, cross-entropy and CTC losses are not directly linked to error rates, and there is no evidence that a teacher with the lowest global loss also provides the lowest error rate. The multi-teacher distillation can, therefore, benefit from the introduction of the ER metric to the training procedure. Hence, we introduce three different weighting strategies: \textit{``Weighted''}, \textit{``Top-1''} and \textit{``Top-k''}. \\

\noindent\textbf{\textit{``Weighted"} Strategy}: A weight is assigned to all the teachers with respect to their average error rates on the current mini-batch, such that the weights dynamically change during training. More precisely, $w_m$ from Eq.\ref{eq9} is computed as the \textit{softmax} distribution from the ER of the current mini-batch:

\begin{equation}
w_m = \frac{\exp{(1 - er_m)}}{\sum_{m=1}^M \exp{(1 - er_m)}}.
\label{eq12}
\end{equation}

In particular, we propose to extend $\mathcal{L}_{CTC-KD}$ to multi teacher distillation by replacing the $N$-best hypothesis with the number of teachers $M$. For instance, in \cite{takashima2018investigation} a beam of size $N$ is necessary to generate enough diverse sentences from a single teacher. In our approach, we replace the beam-search with the $M$ $1$-best sentences generated by $M$ teachers as:

\begin{equation}
\small \mathcal{L}_{CTC-KD} = - \sum_S \sum_{m=1}^{M} p_{tea}(\hat{\textbf{y}}_m|\textbf{h}^e_{tea}) \log p_{st}(\hat{\textbf{y}}_m|\textbf{h}^e_{st}),
\label{eq10}
\end{equation}
with $p_{tea}(\hat{\textbf{y}}_m|\textbf{h}^e_{tea}) = w_m$. This method is extended to $\mathcal{L}_{CE-KD}$ in the same manner but at the frame-level. Nevertheless, with this strategy, the worst teacher would still impact negatively the training even though it has a lowest weight, and the variation of ERs in one mini-batch could be large enough to not reflect properly the quality of a teacher.\\

\noindent
\textbf{\textit{``Top-1"} Strategy}: The best teacher (\textit{i.e.} lowest ER) for each sentence is used during distillation. Similarly to the previous strategy $\mathcal{L}_{CE-KD}$ can dynamically be computed following Eq. \ref{eq4} during training. This approach slightly reduces the computational complexity, but also suffers from a lack of diversity. Indeed, the same teacher will always be picked for a specific sentence from one epoch to an other one. To mitigate this issue, a third strategy is introduced.\\

\noindent
\textbf{\textit{``Top-K"} Strategy}: This method proposes to consider all the teachers that obtain identical error rates at the sentence level as equal candidates for distillation. In particular, identical ER do not necessarily mean that posterior probabilities are also equivalent as different word-level mistakes could be observed. Thus, Eq. \ref{eq4} integrates the $K$-best teachers as:

\begin{equation}
\small \mathcal{L}_{CE-KD} = - \sum_S \sum_{h^d_1}^{h^d_{T_y}} \sum_{k=1}^K \sum_{v\in V} \frac{1}{K} p_{tea}(v|\textbf{h}^d_{tea}) \log p_{st}(v|\textbf{h}^d_{st}).
\label{eq13}
\end{equation}

Finally, the global losses $\mathcal{L}_{KD}$ and $\mathcal{L}_{total}$ are computed with the new $\mathcal{L}_{CE-KD}$ and $\mathcal{L}_{CTC-KD}$ based on Eq. \ref{eq7} and Eq. \ref{eq8} respectively.

During all our experiments, the $\mathcal{L}_{CTC-KD}$ only benefits from the \textit{``Weighted''} strategy while the others are investigated for $\mathcal{L}_{CE-KD}$. As a matter of fact, and following \cite{SB2021}, the CTC decoder is only used to help the attentional decoder learing a correct alignment at training time, and is completely discarded at testing time, lowering the impact and the interest of various KD strategies for this part. 

\begin{table*}
	\caption{Example of the different teacher models used to compose the ensemble on the TIMIT dataset. ``\textit{RC}'' is the number of repeated convolutional blocks, ``\textit{RNN}'' is the type of decoder RNN, ``\textit{neurons}'' is the size of the hidden dimension in the encoder, ``\textit{layers}'' is the number of RNN encoder layers, ``\textit{dropout}'' is the dropout rate, and ``\textit{data aug.}'' represents whether data augmentation (Y) is applied or not (N). The selected teacher to be used as the student is in bold. ``\textit{PER}'' is Phoneme Error Rate.}
	\label{tab:tb1}
	\centering
	\vspace{0.2cm}
	\scalebox{0.8}{
	\begin{tabular}{ccccccccc}
		\toprule
 		RC & RNN & neurons  & layers & dropout & data aug. & batch size & PER valid set \% & PER test set \%  \\
		\midrule
		2&GRU&512 &4 &0.15&Y&8&\textbf{12.38}&13.94 \\
		2&GRU&512 &4 &0.3&N&16&13.51&14.61 \\
		2&GRU&512 &4 &0.3&Y&16&13.36&14.17 \\
		2&LSTM&512 &5 &0.2&N&8&12.64&14.31 \\
		2&GRU&512 &4 &0.3&N&8&12.87&14.32 \\
		2&LSTM&320 &4 &0.3&N&8&14.56&15.61 \\
		1&LSTM&320 &4 &0.3&N&8&15.31&16.81 \\
		2&GRU&640 &4 &0.15&N&8&13.44&15.15  \\
		2&LSTM&512 &5 &0.3&N&8&12.65&14.36 \\
		2&GRU&512 &4 &0.15&N&8&13.27&15.20 \\
		\bottomrule
		
	\end{tabular}
	}
	
\end{table*}

\section{Experiments}
\label{sec:results}

The multi-teacher knowledge distillation approach for joint CTC-attention E2E ASR systems (Sec. \ref{sec:models}) and the proposed distillation strategies are investigated and discussed (Sec. \ref{subsec:results}) under different training strategies on four different datasets of speech recognition (Sec. \ref{subsec:datasets}): TIMIT, Librispeech, Common Voice French and Italian. Finally, we provide an analysis of the distillation procedure (Sec. \ref{subsec:analysis}) to provide insights on the observed performance gain. 

\subsection{Speech recognition datasets}
\label{subsec:datasets}

Four datasets of different sizes, languages and complexities are proposed to validate our multiteacher KD approach.\\

\noindent\textbf{TIMIT \cite{garofolo1993darpa}}. Consists of the standard 462-speaker training set, a 50-speakers development set and a core test set of $192$ sentences for a total of $5$ hours of clean speech. During the experiments, the SA records of the training set are removed and the development set is used for tuning.\\

\noindent\textbf{Common Voice \cite{ardila2019common} (version 6.1)}. Utterances are obtained from volunteers recording sentences all around the world, and in different languages, from smartphones, computers, tablets, etc. The French set contains a total of $328K$ utterances ($475$ hours). The train set consists of $4190$ speakers ($425.5$ hours of speech), while the validation and test sets contain around $24$ hours of speech. The Italian set, on the other hand, contains $89$, $21$ and $22$ hours of Italian training ($58K$ utterances), validation ($13K$ utterances) and test ($13K$ utterances) data.\\ 

\noindent\textbf{Librispeech \cite{panayotov2015librispeech}.} We considered the official ``train-clean-100'' subset of the dataset composed with $100$ hours of speech in English. Models are evaluated on the standard ``dev-clean'' and ``test-clean'' sets. 

\subsection{Ensemble of teachers and distillation}
\label{sec:models}

A total of $40$ different teacher architectures have been generated across the four datasets (\textit{i.e.} 10 for each). For readability reasons, Tab. \ref{tab:tb1} only reports the different topologies for the TIMIT experiments. Nevertheless, the entire list of teachers along with their SpeechBrain training scripts are distributed on GitHub \footnote{\url{https://github.com/yan-gao-GY/Muti-teacherKD}}. We also integrated the TIMIT example as an official SpeechBrain recipe to facilitate further investigations.\\ 

\noindent\textbf{Teacher models.} All teachers share the CRDNN encoder first introduced in \cite{SB2021}. Hence, $N$ repeated convolutional (RC) blocks composed with two $2D$ convolutional layers, a layer normalisation, and a $1D$ frequency pooling operation are followed with a bidirectional RNN (\textit{e.g.} LiGRU, LSTM or GRU). The recurrent network is finally connected to two dense layers with a batch normalisation operation in between. The CTC decoder is a simple linear layer, while the attentional one is a location-aware LSTM / GRU.  Different teachers are obtained by varying all the possible hyperparameters including numbers of neurons per layer, type of RNN, dropout rate, learning rate, attention dimension, etc. Examples for TIMIT are reported in Tab. \ref{tab:tb1} and exact details are given in the configuration files on GitHub and SpeechBrain. Models are fed with $80$ dimensional Mel filter banks with or without data augmentation (SpecAugment). Teachers are trained during $50$, $30$, $50$, and $25$ epochs for TIMIT, Librispeech, CV French and CV Italian respectively, following a validation-based learning rate annealing that ensures an optimal convergence for all models (\textit{i.e.} no more performance can be obtained from longer training).\\
 
\noindent\textbf{Student selection and distillation.} The student model architecture is based on the best performing teacher from the ensemble. Therefore, we propose to pick the best teacher with respect to the best error rate on the validation set of the different datasets. Then, the selected model is trained with KD following the four training strategies detailed in Sec. \ref{sec:strat} and compared to some other baselines (Sec. \ref{subsec:results}). For the initialization scheme, we propose to start from the pre-trained teacher neural parameters except for the last layer that is re-initialized randomly. Then, we fine-tune the whole architecture. More precisely, KD is applied for $50$, $20$, $40$ and $20$ epochs for TIMIT, Librispeech, CV French and CV Italian respectively with SGD and with a validation-based learning rate annealing. For fair comparison, we also fine-tuned the best teacher models  and did not observe any noticeable gain.   

\begin{table}
	\caption{Results observed with the different KD strategies and expressed in term of percentage of Phoneme Error Rate for TIMIT and Word Error Rate for Librispeech and Common Voice (\textit{i.e.} lower is better). Models are evaluated on the test with respect to the best validation performance. ``\textit{Prev. SOTA}'' are state-of-the-art ER obtained from the literature w.r.t the closest experimental setup (\textit{i.e.} E2E ASR without language models).  ``\textit{Original}'' results give the ER obtained with teacher models selected to become students prior to distillation. ``\textit{Average}'' is the baseline multi-teacher KD strategy detailed in Eq. \ref{eq9}. ``\textit{Top-1}'', ``\textit{Top-k}'' and ``\textit{Weighted}'' are the error rate based KD strategies introduced in Sec. \ref{sec:strat}. \\ }
    \label{tab:tb2}
	\centering
	\scalebox{0.87}{
	\begin{tabular}{lcccc}
		\toprule
		& TIMIT & CV Italian & CV French & Libri-100h \\
		\midrule
		Prev. SOTA & 13.8 \cite{ravanelli2019pytorch} & 16.6 \cite{SB2021} & 17.7 \cite{SB2021} & 14.7 \cite{Luscher2019} \\
		\midrule
		Original & 13.94 & 16.54  & 17.73  & 10.61  \\
		Average & 14.58 & 16.86 & 17.84 & 10.86 \\
		Top-1 & 13.15 & 15.71 & \textbf{17.04} & \textbf{10.09} \\
		Top-k & 13.13 & \textbf{15.57} & 17.15  & 10.12 \\
		Weighted & \textbf{13.11} & 15.69 & 17.16 & 10.13 \\
		\bottomrule
	\end{tabular}
	}
\end{table}

\subsection{Speech recognition results}
\label{subsec:results}

Tab. \ref{tab:tb2} shows the Phoneme and Word Error Rates (PER / WER) of the tested student-teacher strategies on the TIMIT, Librispeech and Common Voice test sets. It is important to note that results are obtained with respect to the best validation performance (\textit{i.e.} not tuned on the test sets). We compare our proposed strategies with two different baselines: the literature, and an average weighted distillation strategy. Error rates are reported without further rescoring with a language model or additional self-supervised or unsupervised methods to enable a fair comparison of the acoustic models performance. Literature baselines report previous SOTA results with an hybrid DNN-HMM ASR system for TIMIT \cite{ravanelli2019pytorch}, and two different E2E encoder-decoder with attention systems for Common Voice \cite{SB2021} and Librispeech \cite{Luscher2019}. Best teachers have also been fine-tuned without any KD for more epochs for fair comparison, but failed at improving the performance observed prior to fine-tuning. 

\begin{figure*}[!ht]
    \centering
    \includegraphics[width=360pt]{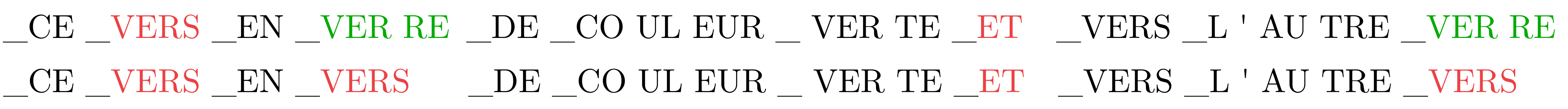}
    \caption{Output tokens (\textit{i.e.} BPE) obtained on a French recording with the best (top) and worst (bottom) teachers. The given sentence contains $5$ occurences of $3$ different homophones (i.e. \textit{``vers''}, \textit{``verre''}, \textit{``vert''}). While the worst teacher is grammatically wrong for $3$ out of $5$, produced tokens remain acoustically correct thus acting as an informed label smoothing during distillation. The original sentence is: \textit{``ce verre en verre de couleur verte est vers l'autre verre''}. }
    \label{fig:analysis}
\end{figure*}

\begin{figure}[!ht]
  \centering
  \includegraphics[width=240pt]{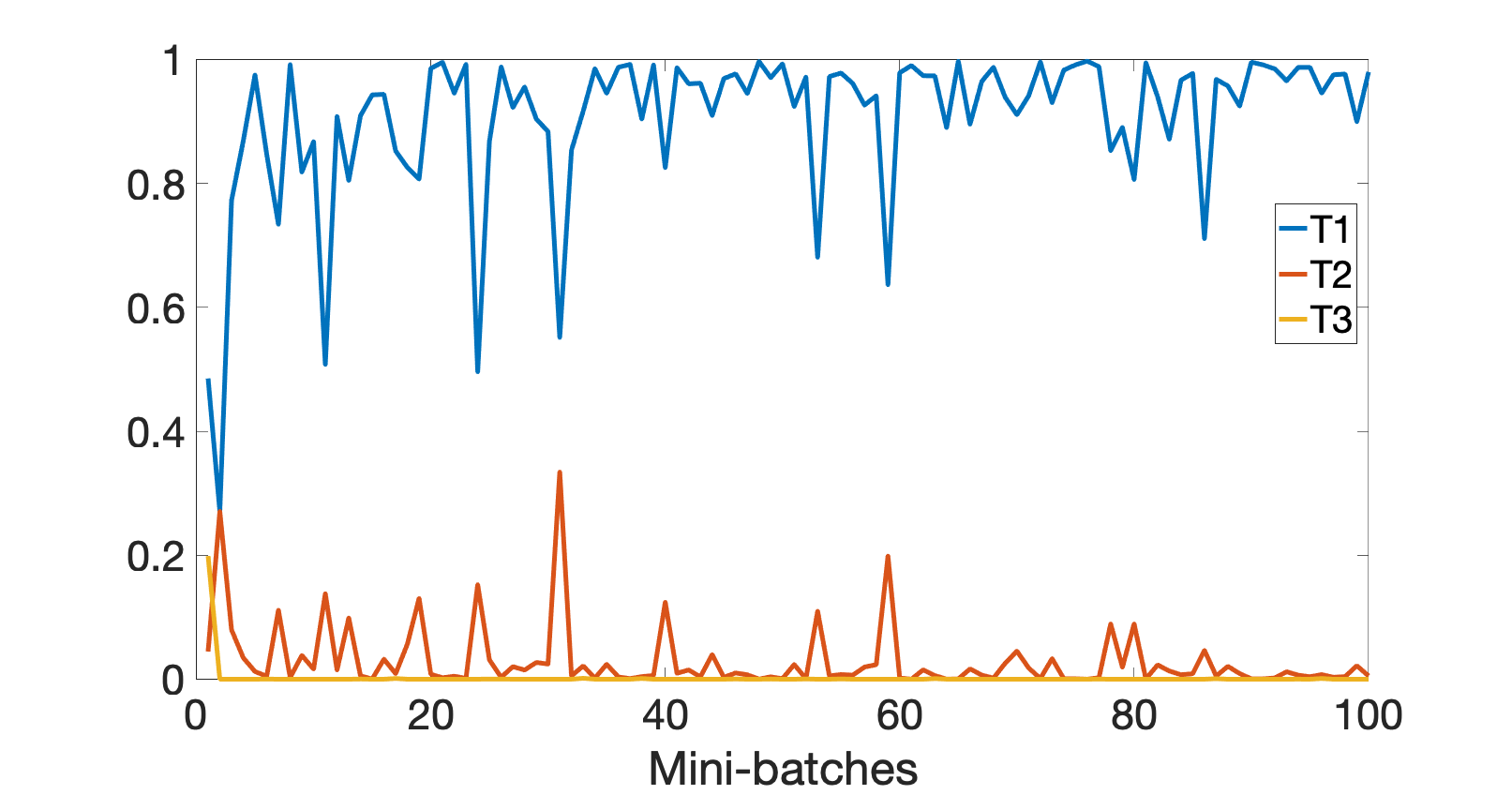}
  \caption{The variation of weights assigned to the first three teachers (Tab. \ref{tab:tb1}) for the CE loss during the first $100$ mini-batches of training on TIMIT with the \textit{``Weighted''} strategy.}
  \label{fig:weight}
\end{figure}

First, it is worth noticing that the proposed error-based distillation strategies obtain SOTA results in the four considered tasks with a PER of $13.11$\% on TIMIT and WER of $17.04$\%, $15.57$\% and $10.09$\% on Common Voice French, Italian and Librispeech respectively, demonstrating the effectiveness of the approach. Interestingly however, the best ER-based stategy is not the same across all tasks. This may be explained by the very nature of the considered dataset. For instance, the ``\textit{Weighted}'' strategy works better than the others on TIMIT while it is the ``\textit{Top-k}'' on CV Italian. TIMIT is an extremely clean dataset, while CV Italian is the exact opposite. Hence, assigning weights to poor teachers with the ``\textit{Weighted}'' strategy on TIMIT is not as negatively impactful as for CV Italian where WER can easily be higher than $100$\%. 

Then, the ``\textit{Average}'' baseline consistently fails at outperforming the original teacher model, empirically demonstrating and validating the importance of teacher weight assignment strategy. This method gives to a poor teacher the same emphasis as a good one. 
Indeed, students obtained from KD with ``\textit{Average}'' degrade the performance on the four datasets compared to the original best teachers.

It is clear from Tab. \ref{tab:tb2} that the three proposed ER-based strategy are viable as they provide competitive performances across most datasets. For example, on Librispeech, ``\textit{Top-k}'' and ``\textit{Weighted}'' are only separated by $0.04$\% of WER. This demonstrates that error rate may be used as a robust proxy to be optimised instead of the loss as the variation in performance observed from different ER strategies remains low. On average, and over the four tasks, error rate based strategies improved the performance in WER by $0.74$ over the best teachers. While such a gain might appear as not being enormous, it is worth underlining that this approach is recycling already trained models instead of simply discarding them and throwing away compute time and energy. During hyper parameter search these additional models could be used collectively to arrive at the final model. 

\subsection{Distillation Analysis}
\label{subsec:analysis}

It is necessary to further investigate the effect of KD on an ensemble of teachers to obtain insights on the observed performance gain. To approach this, we propose two steps: a. empirically illustrate the impact of KD outside of the selected distillation strategy; b. analyse the behavior of the teacher selection within the different ER-based strategies on TIMIT.

Fig. \ref{fig:analysis} illustrates the output tokens obtained with the attentional decoder of the best (top) and worst (bottom) teachers prior to distillation and originally trained on Common Voice French. The given utterance has been recorded in a clean environment and with a consumer-range microphone. We decided to voluntarily select a sentence containing five occurrences of three different French homophones (i.e. \textit{``vers''}, \textit{``verre''}, \textit{``vert''}). From Fig. \ref{fig:analysis}, it becomes clear that KD acts as an informed label smoothing, as bad teachers might provide grammatically wrong outputs that remain acoustically plausible. For instance, the worst teacher (bottom) gives the homophone \textit{``\_VERS''} twice instead of \textit{``\_VER RE''}. Despite being wrong, this teacher still distill the acoustic information giving higher probabilities to words that \textit{``sound the same''} instead of random ones. 

Then, Fig. \ref{fig:weight} and Tab. \ref{tab:tb3} provide the needed information to analyse the impact of the ER-based weighting strategy on the teacher selection process during the TIMIT knowledge distillation stage. For instance, Tab. \ref{tab:tb3} reports the cumulative number of selection for the ten teachers under the \textit{``Top-1''} and \textit{``Top-k''} strategies (same order as in Tab. \ref{tab:tb1}). As expected, good teachers are selected more frequently than bad ones (e.g. T1 is the best performing teacher and is the most selected one with all strategies). Then, it is also confirmed that \textit{``Top-k''} enables a more uniform selection among the teachers. For instance, teacher 8 goes from $1$ selection during \textit{``Top-1''} to $61$ with \textit{``Top-k''}, thus increasing the diversity of labels. Finally, Fig. \ref{fig:weight} depicts the variation of the weights associated to the three first teachers during the first $100$ mini-batches with the \textit{``Weighted''} strategy. Similarly to the others methods, T1 has a far higher average weight than the others. Nevertheless, by crossing with Tab. \ref{tab:tb3} it becomes clear that we will still always associate a weight to bad teachers that are most of the time ignored by the others strategies, thus explaining worse results with noisy datasets.

\begin{table}[!t]
	\caption{Cumulative number of selection for each teacher (Tab. \ref{tab:tb1}) under \textit{``Top-1''} and \textit{``Top-k''} distillation strategies.}
	\vspace{0.2cm}
    \label{tab:tb3}
	\centering
	\scalebox{0.75}{
	\begin{tabular}{lccccccccccccccc}
		\toprule
		  & T1 & T2 & T3 & T4 & T5 & T6 & T7 & T8 & T9 & T10  \\
		\midrule
		 Top-1 & 2361 & 133 & 53 & 12 & 14 & 324 & 17 & 1 & 502 & 280   \\
		 Top-k & 3668 & 775 & 270 & 279 & 309 & 791 & 239 & 61 & 600 & 280  \\
		\bottomrule
	\end{tabular}
	}

\end{table}

\section{Conclusion}
This paper introduces error rate based knowledge distillation from ensembles of teachers for joint ctc-attention end-to-end ASR systems with the definition of three different distillation strategies taking into account sequence-level error rates. The conducted experiments on four different datasets have highlighted promising performance improvements achieved under these strategies with state-of-the-art error rates achieved across all tasks. These findings enable the community to benefit from multiple trained models to further improve the error rate performance instead of discarding them, thus minimizing the loss of compute time.

\bibliographystyle{IEEEbib}
\bibliography{strings,refs}

\end{document}